%% file: root.tex
\def\year{2022}\relax
\documentclass[letterpaper]{article} 
\usepackage{aaai22}  
\usepackage{times}  
\usepackage{helvet}  
\usepackage{courier}  
\usepackage[hyphens]{url}  
\usepackage{graphicx} 
\usepackage{natbib}  
\usepackage{caption} 
\DeclareCaptionStyle{ruled}{labelfont=normalfont,labelsep=colon,strut=off} 
\usepackage{algorithm}

\usepackage{newfloat}
\usepackage{listings}
\floatstyle{ruled}
\newfloat{listing}{tb}{lst}{}
\floatname{listing}{Listing}
\usepackage{multirow}
\usepackage{lscape} 
\usepackage{tabularx} 
\usepackage{graphicx} 
\usepackage{comment}
\usepackage{color}
\usepackage{algorithm}
\usepackage[noend]{algpseudocode} 
\usepackage{amssymb,amsmath}
\newtheorem{theorem}{Theorem}
\newtheorem{lemma}{Lemma}
\newtheorem{definition}{Definition}
\newtheorem{corollary}{Corollary}
\newtheorem{proof}{Sketch Proof}

\title{
	Enhanced Multi-Objective A* Using Balanced Binary Search Trees
}

\author{
    Zhongqiang Ren$^{1}$, Richard Zhan$^{1}$, Sivakumar Rathinam$^{2}$, Maxim Likhachev$^{1}$ and Howie Choset$^{1}$
}
\affiliations{
    \textsuperscript{\rm 1}Carnegie Mellon University, 5000 Forbes Ave., Pittsburgh, PA 15213, USA\\
    \textsuperscript{\rm 2}Texas A\&M University, College Station, TX 77843-3123.\\
    \{zhongqir, rzhan, mlikhach, choset\}@andrew.cmu.edu, srathinam@tamu.edu
}

\begin{document}
	
	\maketitle

	 		\thispagestyle{plain}
	 		\pagestyle{plain}
	 		\pagenumbering{arabic}

	\begin{abstract}
		\input{abstract}
	\end{abstract}
	
	\section{Introduction}\label{sec:intro}
	\input{intro}
	

	\section{Problem Description}
	\input{problem}

	\section{Preliminaries}

\input{method}

	\section{Enhanced Multi-Objective A*}
	\input{checkUpdate}
	
	\section{Tri-Objective A* (TOA*)}
	\input{toastar}

	\section{Analysis of EMOA*}\label{sec:analysis}
	\input{analysis}

	\section{Numerical Results}\label{sec:result}
	\input{result}

	\section{Other Related Work}
	\input{related}

	\section{Conclusion}
	\input{conclude}

\section*{Acknowledgment}
This material is based upon work supported by the National Science Foundation under Grant No. 2120219 and 2120529. Any opinions, findings, and conclusions or recommendations expressed in this material are those of the author(s) and do not necessarily reflect the views of the National Science Foundation.
	
	\newpage

	\bibliography{ref}
	
	\newpage
	\input{appendix}

\end{document}

%% file: abstract.tex
This work addresses a Multi-Objective Shortest Path Problem (MO-SPP) on a graph where the goal is to find a set of Pareto-optimal solutions from a start node to a destination in the graph. A family of approaches based on MOA* have been developed to solve MO-SPP in the literature. Typically, these approaches maintain a ``frontier'' set at each node during the search process to keep track of the non-dominated, partial paths to reach that node. This search process becomes computationally expensive when the number of objectives increases as the number of Pareto-optimal solutions becomes large. In this work, we introduce a new method to efficiently maintain these frontiers for multiple objectives by incrementally constructing balanced binary search trees  within the MOA* search framework. We first show that our approach correctly finds the Pareto-optimal front, and then provide extensive simulation results for problems with three, four and five objectives to show that our method runs faster than existing techniques by up to an order of magnitude.



%% file: intro.tex
Given a graph with non-negative scalar edge costs, the well-known shortest path problem (SPP) requires computing a minimum-cost path from the given start node to a destination node in the graph.
This work considers the so-called Multi-Objective Shortest Path Problem (MO-SPP)~\cite{loui1983optimal,moastar,mandow2008multiobjective}, which generalizes SPP by associating each edge with a non-negative cost vector (of constant length), where each component of the vector corresponds to an objective to be minimized.
MO-SPP arises in many applications including hazardous material transportation~\cite{erkut2007hazardous}, robot design~\cite{xu2021multi}, and airport departure runway scheduling~\cite{montoya2013multiobjective}.

In the presence of multiple conflicting objectives, in general, no (single) feasible path can simultaneously optimize all the objectives. Therefore, the goal of MO-SPP is to find a Pareto-optimal set (of solution paths), whose cost vectors form the so-called Pareto-optimal front.
A path is Pareto-optimal (also called non-dominated) if no objective of the path can be improved without deteriorating at least one of the other objectives.
MO-SPP is computationally hard, even for two objectives~\cite{hansen1980bicriterion}.

To solve MO-SPP, several multi-objective A* (MOA*)-like planners~\cite{moastar,mandow2008multiobjective,ulloa2020simple,goldin2021approximate,ahmadi2021bi} have been developed to compute the exact or an approximated Pareto-optimal front.
In MO-SPP, there are in general, multiple non-dominated partial solution paths between the start and any other node in the graph, and MOA* planners memorize, select and expand these non-dominated paths at each node during the search.
When a new partial solution path $\pi$ is found to reach a node $v$, the path $\pi$ needs to be compared with all previously found non-dominated paths that reach $v$ to check for dominance: verify whether the accumulated cost vector along $\pi$ is dominated by the accumulated cost vector of any existing paths.
These dominance checks are computationally expensive, especially when there are many non-dominated paths at a node, as it requires numerous cost vector comparisons~\cite{pulido2015dimensionality}.

Recently, fast dominance check techniques~\cite{pulido2015dimensionality,ulloa2020simple} were developed under the framework of MOA*-like search to expedite these dominance checks.
Among them, the Bi-objective A* (BOA*) \cite{ulloa2020simple} achieves around an order of magnitude speed up compared to the existing MOA*-like search. Recently, BOA* has been further improved in \cite{goldin2021approximate} and \cite{ahmadi2021bi}.
However, BOA* as well as its improved versions are limited to handle two objectives only.
In this work, we provide a new approach to perform dominance checks relatively fast that can handle an arbitrary number of objectives for MOA*-like search.

Specifically, building on the existing fast dominance check techniques~\cite{pulido2015dimensionality,ulloa2020simple}, we develop a new method called Enhanced Multi-Objective A* (EMOA*) that uses a balanced binary search tree (BBST) to store the non-dominated partial solution paths at each node.
The key ideas are: (i) the BBST can be \emph{incrementally} constructed during the MOA* search, which makes it computationally efficient to maintain; (ii) the BBST is organized using the lexicographic order between cost vectors, which can \emph{guide} the dominance checks and expedite the computation; {(iii) the developed BBST-based method is compatible with existing dominance check approaches, which allows EMOA* to also include the existing techniques to speed up the computation of the Pareto-optimal front.}
We also show that the existing BOA*~\cite{ulloa2020simple} is a special case of EMOA* when there are only two objectives.
We analyze the runtime complexity of the proposed method and show that EMOA* can find all cost-unique Pareto-optimal solutions.
To verify our approach, we run massive tests to compare EMOA* with three baselines (NAMOA*-dr and two extensions of BOA*) in various maps with three, four and five objectives. Our results show that EMOA* achieves up to an order of magnitude speed-up compared to all the baselines {on average}, and is particularly advantageous for problem instances that have a large number of Pareto-optimal solutions.

%% file: problem.tex
Let $G=(V,E,\vec{c})$ denote a graph with vertex set $V$ and edge set $E$, where each edge $e\in E$ is associated with a non-negative cost vector $\vec{c}(e) \in (\mathbb{R^+})^{M}$ with $M$ being a positive integer and $\mathbb{R^+}$ being the set of non-negative real numbers.
Let $\pi(v_1,v_\ell)$ denote a path connecting $v_1,v_\ell \in V$ via a sequence of vertices $(v_1,v_2,\dots,v_\ell)$ in $G$, where $v_k$ and $v_{k+1}$ are connected by an edge $(v_k,v_{k+1}) \in E$, for $k=1,2,\dots,\ell-1$.
Let $\vec{g}(\pi(v_1,v_\ell))$ denote the cost vector corresponding to the path $\pi(v_1,v_\ell)$, which is the sum of the cost vectors of all edges present in the path, $i.e.$ $\vec{g}(\pi(v_1,v_\ell)) = \Sigma_{k=1,2,\dots,\ell-1} \vec{c}(v_k,v_{k+1})$.
To compare any two paths, we compare the cost vector associated with them using the dominance relation~\cite{ehrgott2005multicriteria}:
\begin{definition}[Dominance]\label{def:dominance}
	Given two vectors $a$ and $b$ of length $M$, $a$ dominates $b$ (denoted as $a \succeq b$) if and only if $a(m) \leq b(m)$, $\forall m \in \{1,2,\dots,M\}$, and $a(m) < b(m)$, $\exists m\in \{1,2,\dots,M\}$.
\end{definition}
If $a$ does not dominate $b$, this non-dominance is denoted as $a \nsucceq b$. Any two paths $\pi_1(u,v),\pi_2(u,v)$, for two vertices $u,v \in V$, are non-dominated (with respect to each other) if the corresponding cost vectors do not dominate each other.

Let $v_o,v_d$ denote the start and destination vertices respectively. The set of all non-dominated paths between $v_o$ and $v_d$ is called the {\it Pareto-optimal} set.
A maximal subset of the Pareto-optimal set, where any two paths in this subset do not have the same cost vector is called a {\it cost-unique} Pareto-optimal set. This paper considers the problem that aims to compute a cost-unique Pareto-optimal set.

%% file: method.tex
\subsection{Basic Concepts}
Let $l=(v,\vec{g})$ denote a \emph{label}\footnote{To identify a partial solution path, different names such as nodes~\cite{ulloa2020simple}, states~\cite{ren21momstar, ren21mopbd} and labels~\cite{martins1984multicriteria,sanders2013parallel}, have been used in the multi-objective path planning literature. This work uses ``labels'' to identify partial solution paths.}, which is a tuple of a vertex $v\in V$ and a cost vector $\vec{g}$.
A label represents a partial solution path from $v_o$ to $v$ with cost vector $\vec{g}$.
To simplify notations, given a label $l$, let $v(l), \vec{g}(l)$ denote the vertex and the cost vector contained in label $l$ respectively. A label $l$ is said to be dominated by (or is equal to) another label $l'$ if $v(l)=v(l')$ and $\vec{g}(l) \succeq \vec{g}(l')$ (or $\vec{g}(l)=\vec{g}(l')$).

Let $\vec{h}(v), v \in V$ denote a \emph{consistent} heuristic vector of vertex $v$ that satisfies $\vec{h}(v) \leq \vec{h}(u) + \vec{c}(u,v), \forall u,v \in V$.
Additionally, let $\vec{f}(l) := \vec{g}(l) + \vec{h}(v(l))$, and let OPEN denote a priority queue of labels, where labels are prioritized by their corresponding $\vec{f}$-vectors in \emph{lexicographic} order.

Finally, let $\alpha(u), u \in V$ denote the \emph{frontier} set \emph{at} vertex $u$, which stores all non-dominated labels $l$ \emph{at} vertex $u$ (i.e. $v(l) = u$).
Intuitively, each label $l \in \alpha(u), u \in V$ identifies a non-dominated (partial solution) path from $v_o$ to $u$.
See Fig.~\ref{fig:emoa_check} (a) for an illustration.
For presentation purposes, we also refer to $\alpha(v_d)$ as $\mathcal{S}$, the solution set, which is the frontier set at the destination vertex, and each label in $\mathcal{S}$ identifies a {cost-unique} Pareto-optimal solution (path). 


\subsection{Search Framework}
To begin with, we reformulate BOA*~\cite{ulloa2020simple} as a general search framework as shown in Alg.~\ref{alg:emoa}, and explain the running process.
We then provide a technical review of the existing algorithms NAMOA*-dr~\cite{pulido2015dimensionality} and BOA*~\cite{ulloa2020simple}, and discuss how our EMOA* differs from them.

\begin{algorithm}[htbp]
	\caption{Search Framework}\label{alg:emoa}
	\begin{algorithmic}[1]
		\State{$l_o \gets (v_o, \vec{0})$}
		\State{Add $l_o$ to OPEN}
		\State{$\alpha(v)\gets \emptyset, \forall v \in V$}
		\While{OPEN $\neq \emptyset$}
		\State{$l \gets $ OPEN.pop()}
		\If{\textit{FrontierCheck}($l$) \textbf{or} \textit{SolutionCheck}($l$)}
		\State{\textbf{continue}}\Comment{Current iteration ends}
		\EndIf
		\State{\textit{UpdateFrontier}($l$)}
		\If{$v(l) = v_d$}
		\State{\textbf{continue}}\Comment{Current iteration ends}
		\EndIf
		\ForAll{$v' \in$ GetSuccessors($v(l)$)}
		\State{$l' \gets (v', \vec{g}(l) + \vec{c}(v,v'))$, $parent(l')\gets l$}
		\State{$\vec{f}(l') \gets \vec{g}(l') + \vec{h}(v(l'))$}
		\If{\textit{FrontierCheck}($l'$) \textbf{or} \textit{SolutionCheck}($l'$)}
		\State{\textbf{continue}}\Comment{Move to the next successor.}
		\EndIf
		\State{Add $l'$ to OPEN}
		\EndFor
		\EndWhile
		\State{\textbf{return} $\alpha(v_d)$}\Comment{$\alpha(v_d)$ is also referred to as $\mathcal{S}$}
	\end{algorithmic}
\end{algorithm}

As shown in Alg.~\ref{alg:emoa}, to initialize (lines 1-3), an initial label $l_o=(v_o,\vec{0})$ is created, and is added to OPEN.
Additionally, the frontier sets at all vertices are initialized to be empty sets.
In each search iteration (lines 5-16), the label with the lexicographically minimum $\vec{f}$-vector is popped from OPEN and is denoted as $l$ in Alg.~\ref{alg:emoa}.
This label $l$ is then (line 6) checked for dominance via the following two procedures:
\begin{itemize}
	\item First, $l$ is compared with labels in $\alpha(v(l))$ in procedure \textit{FrontierCheck} to verify if there exists a label in $\alpha(v(l))$ that dominates (or is equal to) $l$.
	\item Second, $l$ is compared with labels in $\mathcal{S}$ in procedure \textit{SolutionCheck} to verify if there exists a label $l^*$ in $\mathcal{S}$ such that $g(l^*)$ dominates (or is equal to) $\vec{f}(l)$. Note that $\vec{f}(l^*)=\vec{g}(l^*)$ as $\vec{h}(v(l^*))=\vec{h}(v_d)=\vec{0}$.
\end{itemize}
If $l$ is dominated in either \textit{FrontierCheck} or \textit{SolutionCheck}, $l$ is discarded and the current search iteration ends, because $l$ cannot lead to a cost-unique Pareto-optimal solution.
Otherwise ($i.e.$ $l$ is non-dominated in both \textit{FrontierCheck} and \textit{SolutionCheck}),
$l$ is used to update the frontier set $\alpha(v(l))$ (line 8): procedure \textit{UpdateFrontier} first removes all the existing labels in $\alpha(v(l))$ that are dominated by $l$, and then adds $l$ into $\alpha(v(l))$.
Note that this includes the case when $v(l) = v_d$, where $\alpha(v_d)$ ($i.e.$ $\mathcal{S}$) is updated, which means a solution path from $v_o$ to $v_d$ is found.
After \textit{UpdateFrontier}, label $l$ is verified whether $v(l)=v_d$ (line 9).
If $v(l)=v_d$, the current search iteration ends (line 10); Otherwise, $l$ is expanded, as explained in the next paragraph.

To expand a label $l$ ($i.e.$ to expand the partial solution path represented by label $l$), for each successor vertex $v'$ of $v(l)$, a new label $l'=(v',\vec{g}(l) + \vec{c}(v,v'))$ is generated, which represents a new path from $v_o$ to $v'$ via $v(l)$ by extending $l$ ($i.e.$ the path represented by $l$).
The parent pointer $parent(l')$ is set to $l$, which helps reconstruct a solution path for each label in $\alpha(v_d)$ after the algorithm terminates.
Then (line 14), \textit{FrontierCheck} and \textit{SolutionCheck} are invoked on label $l'$ to verify if $l'$ is dominated and should be discarded or not.
(Note that the dominance checks are needed at both line 6 and 14, which is explained in the next subsection.)
If $l'$ is non-dominated, $l'$ is added to OPEN for future expansion.

Finally (line 17), the search process terminates when OPEN is empty, and returns $\alpha(v_d)$, a set of labels, each of which represents a cost-unique Pareto-optimal solution path.
Additionally, the cost vectors of labels in $\alpha(v_d)$ form the entire Pareto-optimal front of the given problem instance.


\subsection{Brief Summary of NAMOA*-dr and BOA*}
NAMOA*-dr~\cite{pulido2015dimensionality} is a multi-objective search algorithm that can handle an arbitrary number of objectives.
The main differences between NAMOA*-dr and Alg.~\ref{alg:emoa} can be summarized in the following two points.
{\it First}, unlike Alg.~\ref{alg:emoa}, in NAMOA*-dr, when a new label $l'$ is generated during the expansion, $l'$ will be used for dominance checks against \emph{existing} labels in OPEN and frontier sets\footnote{NAMOA*-dr maintains two frontier sets $G_{op}(v)$ (open) and $G_{cl}(v)$ (closed) at each vertex $v\in V$, and we refer the reader to~\cite{ulloa2020simple,pulido2015dimensionality} for more details.} to remove labels that are dominated by $l$ (which happens between line 15 and 16 in Alg.~\ref{alg:emoa} and is not shown in Alg.~\ref{alg:emoa} for presentation purposes).
These checks are called ``eager checks''~\cite{ulloa2020simple}.
With eager checks, each popped label from OPEN is guaranteed to be non-dominated, and lines 6-8 in Alg.~\ref{alg:emoa} are thus skipped in NAMOA*-dr.
{\it Second}, a key idea in NAMOA*-dr is that, with (i) consistent heuristics and (ii) an OPEN list where labels are lexicographically prioritized, the first component of the cost vectors can be ignored in some of the dominance checks.
This idea is referred to as the ``dimensionality reduction'' \cite{pulido2015dimensionality}, which helps in speeding up the dominance checks in NAMOA*-dr.

BOA*~\cite{ulloa2020simple} leverages the aforementioned dimensionality reduction, and introduces the idea of ``lazy checks'', which avoids the eager checks as in NAMOA*-dr.
Specifically, BOA* follows the search process as shown in Alg.~\ref{alg:emoa}, where lines 6-8 in Alg.~\ref{alg:emoa} help defer the eager checks related to a label $l$ until $l$ is going to be expanded.
With the help of this lazy check technique, BOA* guarantees that all dominance checks can be performed in constant time for a bi-objective problem which leads to speeding up the overall search process.
Specifically, the three key procedures \emph{FrontierCheck}, \emph{SolutionCheck} and \emph{UpdateFrontier} as shown in Alg.~\ref{alg:emoa} can be conducted in constant time in BOA*. 

Our approach EMOA* also follows the same framework as shown in Alg.~\ref{alg:emoa}, and inherits the ideas of dimensionality reduction and lazy check.
However, EMOA* realizes the three key procedures by incrementally building balanced binary search trees, which can handle an arbitrary number of objectives (Fig.~\ref{alg:emoa}). This leads to up to an order of magnitude speed-up in comparison with the existing baselines.

%% file: checkUpdate.tex

This section shows how EMOA* realizes
\emph{FrontierCheck}, \emph{SolutionCheck} and \emph{UpdateFrontier} by leveraging balanced binary search trees (BBST).
We begin by introducing a few definitions and then elaborate the BBST-based procedures.

\begin{figure}[t]
	\centering
	\includegraphics[width=0.8\linewidth]{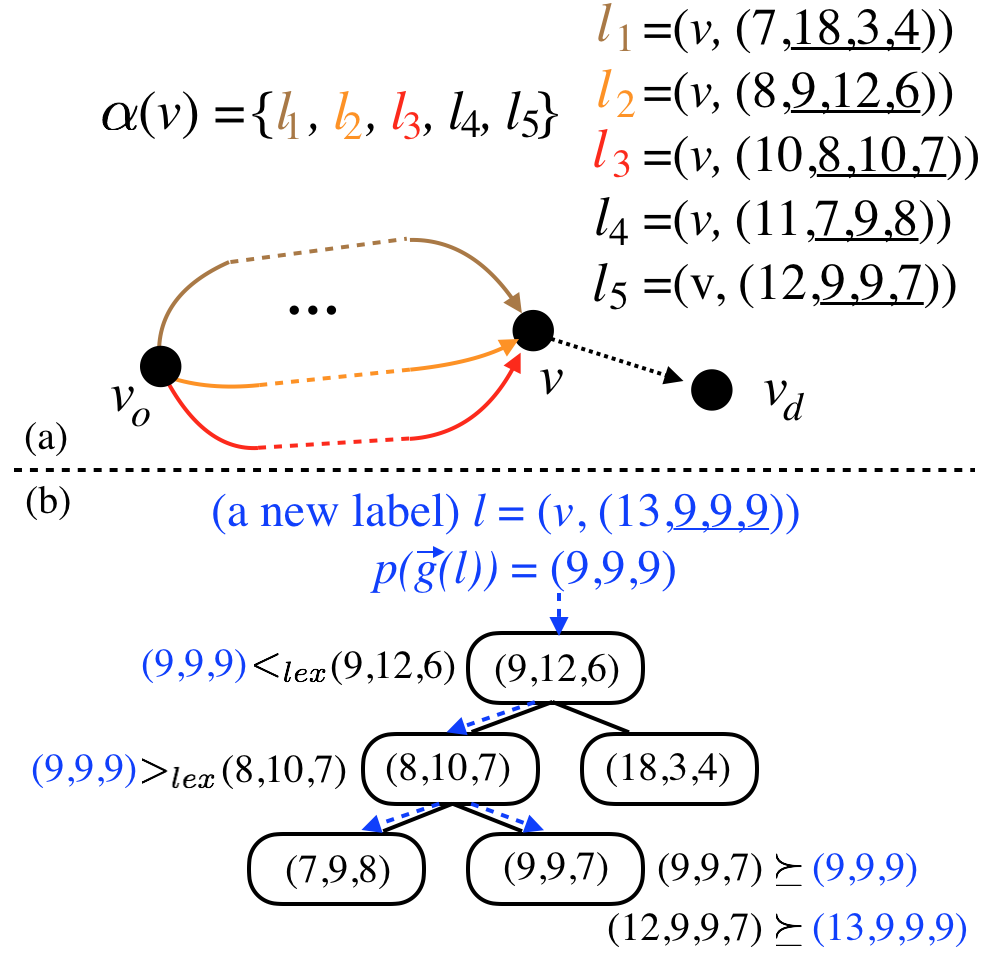}
	\caption{
		Fig. (a) shows the frontier set $\alpha(v)$ at some vertex $v$. There are five labels in $\alpha(v)$. The underlined three numbers of each $\vec{g}$-vector indicate the corresponding projected vector $p(\vec{g})$.
		Fig. (b) shows the corresponding balanced binary search tree. The keys of the nodes in this tree forms the non-dominated subset of the projected vectors.
		The dashed blue arrows show the sequence of tree nodes that are traversed when running the \emph{FrontierCheck} procedure (Alg.~\ref{alg:check}).
		The projected vector $(9,9,7)$ in the tree dominates the input vector $(9,9,9)$, which indicates that the new label $l$ (in blue) with $\vec{g}$-vector $(13,9,9,9)$ is dominated and should be discarded in EMOA* search.
	}
	\label{fig:emoa_check}
\end{figure}

\subsection{The Check and Update Problems}

\begin{definition}[Dominance Check (DC) Problem]\label{problem:check}
	Given a set $B$ of $K$-dimensional non-dominated vectors and a new $K$-dimensional vector $b$, the DC problem aims to verify whether there exists a vector $b' \in B$ such that $b' \leq b$ ($i.e.$ $b'$ is component-wise no larger than $b$, which is equivalent to $b' \succeq b$ or $b' = b$).
\end{definition}
\begin{definition}[Non-Dominated Set Update (NSU) Problem]\label{problem:update}
	Given a set $B$ of $K$-dimensional non-dominated vectors and a new $K$-dimensional vector $b$ that is non-dominated by any vectors in $B$, the NSU problem computes $ND(B\bigcup\{b\})$, (i.e. the non-dominated subset of $B\bigcup\{b\}$).
\end{definition}
The relationship between the aforementioned three procedures and these two problems can be described as follows:
\begin{itemize}
	\item In \textit{FrontierCheck}, given a new label $l$ and the frontier set $\alpha(v(l))$, an equivalent DC problem can be generated with input $b=\vec{g}(l)$ and $B=\{\vec{g}(l') | l' \in \alpha(v(l))\}$.
	\item Similarly, in \textit{SolutionCheck}, given a new label $l$ and the frontier set $\alpha(v(l))$, an equivalent DC problem can be generated with $b=\vec{f}(l)$ and $B=\{\vec{g}(l') | l' \in \alpha(v_d)\}$.
	\item Finally, in \textit{UpdateFrontier}, given a new label $l$ and the frontier set $\alpha(v(l))$, an equivalent NSU problem can be generated with $b=\vec{g}(l)$ and $B=\{\vec{g}(l') | l' \in \alpha(v(l))\}$.
\end{itemize}
From now on, we focus on how to quickly solve DC and NSU problems.
Note that, a \emph{baseline} method that solves the DC problem runs a for-loop over each vector $b' \in B$ and check if $b' \leq b$, which takes $O(|B|K)$ time.
A baseline method that solves the NSU problem requires two steps: (i) filter $B$ by removing from $B$ all vectors that are dominated by $b$, and (ii) add $b$ into $B$. Here, a naive method for step (i) runs a for-loop over set $B$ to remove all dominated vectors and takes $O(|B|K)$ time, and step (ii) takes constant time.
Consequently, the overall time complexity is $O(|B|K)$.

\subsection{Balanced Binary Search Trees (BBSTs)}
To efficiently solve the DC and NSU problems, our method leverages the BBST data structure.
As a short review, let $n$ denote a node\footnote{For the rest of this work, for presentation purposes, the term ``vertex'' is associated with the graph $G$ and the term ``node'' is associated with the balanced binary search tree.} within a binary search tree (BST) with the following attributes:
\begin{itemize}
	\item Let $n.height$ denote the height of node $n$, which is the number of edges along the longest downwards path between $n$ and a leaf node. A leaf node has a height of zero.
	\item Let $n.key$ denote the key of $n$, which is a $K$-dimensional vector in this work. To compare two nodes, their keys are compared by \emph{lexicographic} order.
	\item Let $n.left$ and $n.right$ denote the left child and the right child of $n$ respectively, which represent the left sub-tree and the right sub-tree respectively.
	\item We say $n=NULL$ if $n$ does not exist in the BST. For example, if $n$ is a leaf node, then $n.left=NULL$ and $n.right=NULL$.
\end{itemize}
In this work, we limit our focus to the AVL-tree, one of the most famous balanced BSTs.
An AVL-tree has the following property: for any node $n$ within an AVL-tree, let $d(n) := n.left.height - h.right.height$ denote the difference between the height of the left and the right child node, then AVL-tree is called ``balanced'' if $d(n) \in \{-1,0,1\}$.
To maintain balance at insertion or deletion of nodes, an AVL-tree invokes the so-called ``rotation'' operations when $d(n) \leq -2 $ or $d(n) \geq 2$ and the tree is always balanced.
Consequently, given an AVL-tree of size $N$ ($i.e.$ containing $N$ nodes), the height of the root is bounded by $O(log N)$.

\subsection{BBST-Based Check Method}


Given a set $B$ of non-dominated vectors, let $\mathcal{T}_B$ denote an AVL-tree that stores all vectors in $B$ as the keys of tree nodes.
Now, given a new vector $b$, the DC problem can be solved via Alg.~\ref{alg:check}, which traverses the tree recursively while running dominance comparison.

Specifically, Alg.~\ref{alg:check} is invoked with \textit{Check}$(\mathcal{T}_B.root, b)$, where $\mathcal{T}_B.root$ denotes the root of the tree and $b$ is the input vector to be checked.
As a base case (line 2), if the input node $n$ is $NULL$, the algorithm terminates and returns false, which means $b$ is non-dominated.
When the input node is not $NULL$, $b$ is checked for dominance against $n.key$ and returns true if $n.key \leq b$.
Otherwise, the algorithm verifies if $b$ is lexicographically smaller than (denoted as $<_{lex}$) $n.key$.
\begin{itemize}
	\item (Case-1) If $b <_{lex} n.key$, there is no need to traverse the right sub-tree from $n$, since any node in the right sub-tree of $n$ cannot be component-wise no larger than $b$, and the algorithm (recursively) invokes itself to traverse the left sub-tree for dominance checks.
	\item (Case-2) Otherwise ($i.e.$ $b >_{lex} n.key$), the algorithm first invokes itself to traverse the left sub-tree (line 9) and then the right sub-tree (line 11) for dominance checks. Note that, in this case, both child nodes need recursive traversal to ensure correctness.
\end{itemize}

\vspace{-2mm}
\begin{algorithm}[htbp]
	\caption{\textit{Check}($n,b$)}\label{alg:check}
	\begin{algorithmic}[1]
		\State{INPUT: $n$ is a node in an AVL-tree and $b$ is a vector}
		\If{$n = NULL$}
		\State{\textbf{return} false}
		\EndIf
		\If{$n.key \leq b$}
		\State{\textbf{return} true}
		\EndIf
		\If{$b <_{lex} n.key$}
		\State{\textbf{return} \textit{Check}($n.left, b$)}
		\Else \Comment{$i.e.$ $b >_{lex} n.key$}
		\If{\textit{Check}($n.left, b$)}\Comment{Removed in TOA*}
		\State{\textbf{return} true}\Comment{Removed in TOA*}
		\EndIf
		\State{\textbf{return} \textit{Check}($n.right, b$)}
		\EndIf
	\end{algorithmic}
\end{algorithm}

\subsection{BBST-Based Update Method}

Similarly, the NSU problem with input $B$ (in the form of a correpsonding BBST $\mathcal{T}_B$) and a non-dominated vector $b$, can be solved by (i) invoking Alg.~\ref{alg:filter} to remove nodes with dominated keys from the tree $\mathcal{T}_B$ and (ii) insert the input (non-dominated) vector $b$ into the tree.
Here, step (ii) is a regular AVL-tree insertion operation, which takes $O(log|B|)$ time, and we will focus on step (i) in the ensuing paragraphs.

\begin{algorithm}[htbp]
	\caption{\textit{Filter}($n,b$)}\label{alg:filter}
	\begin{algorithmic}[1]
		\State{INPUT: $n$ is a node in an AVL-tree and $b$ is a vector}
		\If{$n = NULL$}
		\State{\textbf{return} $NULL$}
		\EndIf
		\If{$b >_{lex} n.key$}
		\State{$n.right \gets$\textit{Filter}($n.right, b$)}
		\Else
		\State{$n.left \gets$\textit{Filter}($n.left, b$)}
		\State{$n.right \gets$\textit{Filter}($n.right, b$)}
		\EndIf
		\If{$b \succeq n.key$}
		\State{\textbf{return} \textit{AVL-Delete}($n$)}
		\EndIf
		\State{\textit{AVL-balancing}() when needed.}
	\end{algorithmic}
\end{algorithm}

For step (i), Alg.~\ref{alg:filter} is invoked with Filter$(\mathcal{T}_B.root, b)$, where $\mathcal{T}_B.root$ denotes the root of the tree and $b$ is the input non-dominated vector.
As shown in Alg.~\ref{alg:filter}, as a base case (line 2), if the input node is $NULL$, the algorithm terminates and returns a $NULL$.
When the input node $n$ is not $NULL$, the algorithm verifies whether $b>_{lex} n.key$.
\begin{itemize}
	\item (Case-1) If $b >_{lex} n.key$, there is no need to filter the left sub-tree of $n$ (since any node in the left sub-tree of $n$ must be non-dominated by $b$) and the algorithm recursively invokes itself to traverse the right sub-tree for filtering.
	\item (Case-2) Otherwise (i.e. $b <_{lex} n.key$)\footnote{Note that it's impossible to have $b=n.key$: Within the EMOA* algorithm (Alg.~\ref{alg:emoa}), \textit{UpdateFrontier} is always invoked after \textit{FrontierCheck}. if $b=n.key$, \textit{FrontierCheck} removes it and \textit{UpdateFrontier} will not be invoked (line 6-8 in Alg.~\ref{alg:emoa}).}, the algorithm first invokes itself to traverse the left sub-tree (line 9) and then the right sub-tree (line 11) for filtering. Note that, in this case, both child nodes need to be traversed for further dominance check to ensure correctness.
\end{itemize}
At the end (line 9-10), $n.key$ is checked for dominance against $b$. If $n.key$ is dominated, $n$ is removed from the tree. The tree is also processed to ensure that the resulting tree is still balanced (line 11).
These are all common operations related to AVL-trees.
In the worst case, the entire tree is traversed and all nodes in the tree are recursively deleted (from the leaves to the root), which takes $O(|B|K)$ time.

\vspace{1mm}
\noindent\textbf{Remark.} Theoretically, both Alg.~\ref{alg:check} and \ref{alg:filter} runs in $O(|B|K)$ time in the worst case, which is the same as the aforementioned baseline approaches ($i.e.$ running a for-loop over $B$).
However, as shown in the result section, these BBST-based methods can solve the DC and NSU problems much more efficiently in practice.
The intuitive reason behind such efficiency is that, the AVL-tree is organized based on the lexicographic order, which can provide guidance when traversing the tree for dominance checks.
As a result, only a small portion of the tree is traversed.
Finally, note that the method in this section does not put any restriction on $K$.


\subsection{EMOA* with BBST-Based Check and Update}
This section presents how to use the BBST-based algorithms (Alg.~\ref{alg:check},~\ref{alg:filter}) within the framework of Alg.~\ref{alg:emoa}.
Specifically, EMOA* leverages the idea of dimensionality reduction, which can expedite the BBST-based check and update.
EMOA* has the property that, during the search process, the sequence of labels being expanded at the same vertex has non-decreasing $f_1$ values, where $f_1$ represents the first component of the $\vec{f}$-vector of a label.
This property is caused by the fact that heuristics are consistent and all labels are selected from OPEN by lexicographic order of their $\vec{f}$-vectors.
Additionally, since all labels at the same vertex $v$ have the same $\vec{h}$-vector, the sequence of labels being expanded at the same vertex also has non-decreasing $g_1$ values, where $g_1$ represents the first component of the $\vec{g}$.

To simplify presentation, let $p:\mathbb{R}^{M} \rightarrow \mathbb{R}^{M-1}$ denote a \emph{projection function} that removes the first component from the input vector.
During the EMOA* search, when a new label $l$ is generated, for \emph{FrontierCheck}, we only need to do dominance comparison between $p(\vec{g}(l))$ and $p(\vec{g}(l')), \forall l' \in \alpha(v(l))$, instead of comparing $\vec{g}(l)$ with $\vec{g}(l'), \forall l' \in \alpha(v(l))$.
Consequently, in EMOA*, for each vertex $v \in V$, a BBST $\mathcal{T}_B$ as aforementioned is constructed with $B=ND(\{ p(\vec{g}(l')), \forall l' \in \alpha(v) \})$.
In other words, the key of each node in $\mathcal{T}_B$ is a non-dominated projected cost vector of a label in $\alpha(v)$.

To realize \textit{FrontierCheck} for a label $l$ that is extracted from OPEN (line 6 in Alg.~\ref{alg:emoa}), Alg.~\ref{alg:check} is invoked with $b=p(\vec{g}(l))$ and $n$ being the root node of the tree $\mathcal{T}_B$. We provide a toy example for \textit{FrontierCheck} in Fig.~\ref{fig:emoa_check}.
Similarly, for \emph{SolutionFilter} (line 6 in Alg.~\ref{alg:emoa}), Alg.~\ref{alg:check} is invoked with $b=p(\vec{f}(l))$ and $n$ being the root node of the tree $\mathcal{T}_{B'}$ with $B'=ND(\{ p(\vec{g}(l')), \forall l' \in \alpha(v_d) \})$ ($i.e.$ the set of all non-dominated projected vectors of labels in the frontier set at the destination node).
During the search, when a label $l$ is extracted from OPEN and is used to update the frontier set in procedure \textit{UpdateFrontier} (line 8 in Alg.~\ref{alg:emoa}), Alg.~\ref{alg:filter} is first invoked with $\vec{b}=p(\vec{g}(l))$ and $n$ being the root node of the tree $\mathcal{T}_B$ where $B=ND(\{ p(\vec{g}(l')), \forall l' \in \alpha(v(l)) \})$.
Then, $\vec{b}=p(\vec{g}(l))$ is added to $\mathcal{T}_B$.

In summary, to realize procedures \textit{FrontierCheck}, \textit{SolutionCheck} and \textit{UpdateFrontier}, only the projected vectors of labels are needed, instead of the original vectors.

\subsection{Generalization of BOA*}
EMOA* generalizes BOA* in the following sense.
When $M=2$, for any cost vector $\vec{g}$ in a label, the projected vector $p(\vec{g})$ is of length one and is thus a scalar value. In this case, the AVL-tree corresponding to $\alpha(v)$ of any vertex $v\in V$ in EMOA* becomes a singleton tree: a tree with a single root node $\mathcal{T}_B.root$.
The key value of $\mathcal{T}_B.root$ is the minimum value of $g_2(l)$ among all labels $l \in \alpha(v)$, which is the same as the auxiliary variable $g^{min}_2$ introduced at each vertex in BOA*.
Solving a DC problem requires only a scalar comparison between $\mathcal{T}_B.root.key$ and the scalar $p(\vec{g})$, the projected cost vector of the label selected from OPEN in each search iteration.
Clearly, this scalar comparison takes constant time.
Additionally, the \textit{UpdateFrontier} in EMOA* requires simply assigning the scalar $p(\vec{g})$ to $\mathcal{T}_B.root.key$ ($i.e.$ $g^2_{min}$), which also takes constant time.
Therefore, BOA* is a special case of EMOA* when $M=2$.




%% file: toastar.tex
When $M=3$, EMOA* can be further improved to achieve better theoretic runtime complexity when running Alg.~\ref{alg:check}.
We name this improved algorithm TOA* (Tri-Objective A*).
TOA* differs from EMOA* as follows: line 9-10 in Alg.~\ref{alg:check} are removed.
In other words, when $M=3$, each projected vector $b$ as well as the key of all nodes in the tree $\mathcal{T}_B$ have length $(M-1)=2$.
In this case, during the computation of Alg.~\ref{alg:check}, when $b >_{lex} n.key$ ($i.e.$ line 8 in Alg.~\ref{alg:check}), there is no need to further traverse the left sub-tree.
\begin{theorem}
	Given $b$, a two dimensional vector, and an arbitrary node $n$ in $\mathcal{T}_B$, if (i) $n.key \nleq b$ and (ii) $b >_{lex} n.key$, then the key of any nodes in the left sub-tree of $n$ cannot dominate $b$.
\end{theorem}
Please find the proof in the appendix.


In TOA*, the modified Alg.~\ref{alg:check} traverses the tree either to the left sub-tree (when $b<_{lex} n.key$) or to the right sub-tree (when $b>_{lex} n.key$), which leads to a time complexity of $O(log|B|)$ (note that $M$ is a constant here).
We say that TOA* is an improved version of EMOA* with $M=3$ since the theoretic computational complexity is improved.
Finally, we summarize the computational complexity of the \textit{Check} and \textit{Update} procedures in BOA*~\cite{ulloa2020simple} and our algorithms (TOA* and EMOA*) in Table~\ref{tab:complexity}.

\begin{table}[htbp]
	\centering
	\small
	\begin{tabular}{ | l | l | l | l |  }
		\hline
		 & BOA* & TOA* & EMOA* \\
		\hline
		$M$ & $=2$ & $=3$ & $\geq 2$ \\
		\hline
		\textit{Check} & Constant Time & $O(log|B|)$ & $O(|B|(M-1))$ \\
		\hline
		\textit{Update} & Constant Time & $O(|B|)$ & $O(|B|(M-1))$ \\
		\hline
	\end{tabular}
	\caption{Runtime complexity of related methods. BOA* is a special case of EMOA* when $M=2$, and TOA* is an improved version of EMOA* when $M=3$.}
	\label{tab:complexity}
\end{table}

%% file: analysis.tex
To save space, we provide the proofs in the appendix.

\begin{lemma}\label{lem:non_decreasing_f1_successor}
	When expanding a label $l$, each successor label $l'$ has $f_1$ value no smaller than the $f_1$ value of $l$.
\end{lemma}

\begin{lemma}\label{lem:non_decrease_f1}
	During the search of Alg.~\ref{alg:emoa}, the sequence of extracted labels from OPEN has non-decreasing $f_1$ values.
\end{lemma}

\begin{corollary}\label{cor:non_decrease_f1_at_a_vertex}
	During the search, the sequence of extracted and expanded labels at a specific vertex has non-decreasing $f_1$ and $g_1$ values.
\end{corollary}

\begin{corollary}\label{cor:dominance_after_proj}
	During the search, for a label $l$ that is extracted from OPEN, we have $\vec{g}(l')\leq \vec{g}(l), l' \in\alpha(v(l))$ if and only if $p(\vec{g}(l')) \leq p(\vec{g}(l)), l' \in\alpha(v(l))$.
\end{corollary}

\begin{theorem}\label{thm:pareto-optimal}
	EMOA* computes a maximal set of cost-unique Pareto-optimal paths connecting $v_o$ and $v_d$ at termination.
\end{theorem}

%% file: result.tex
\begin{figure*}[tb]
	\centering
	\includegraphics[width=\linewidth]{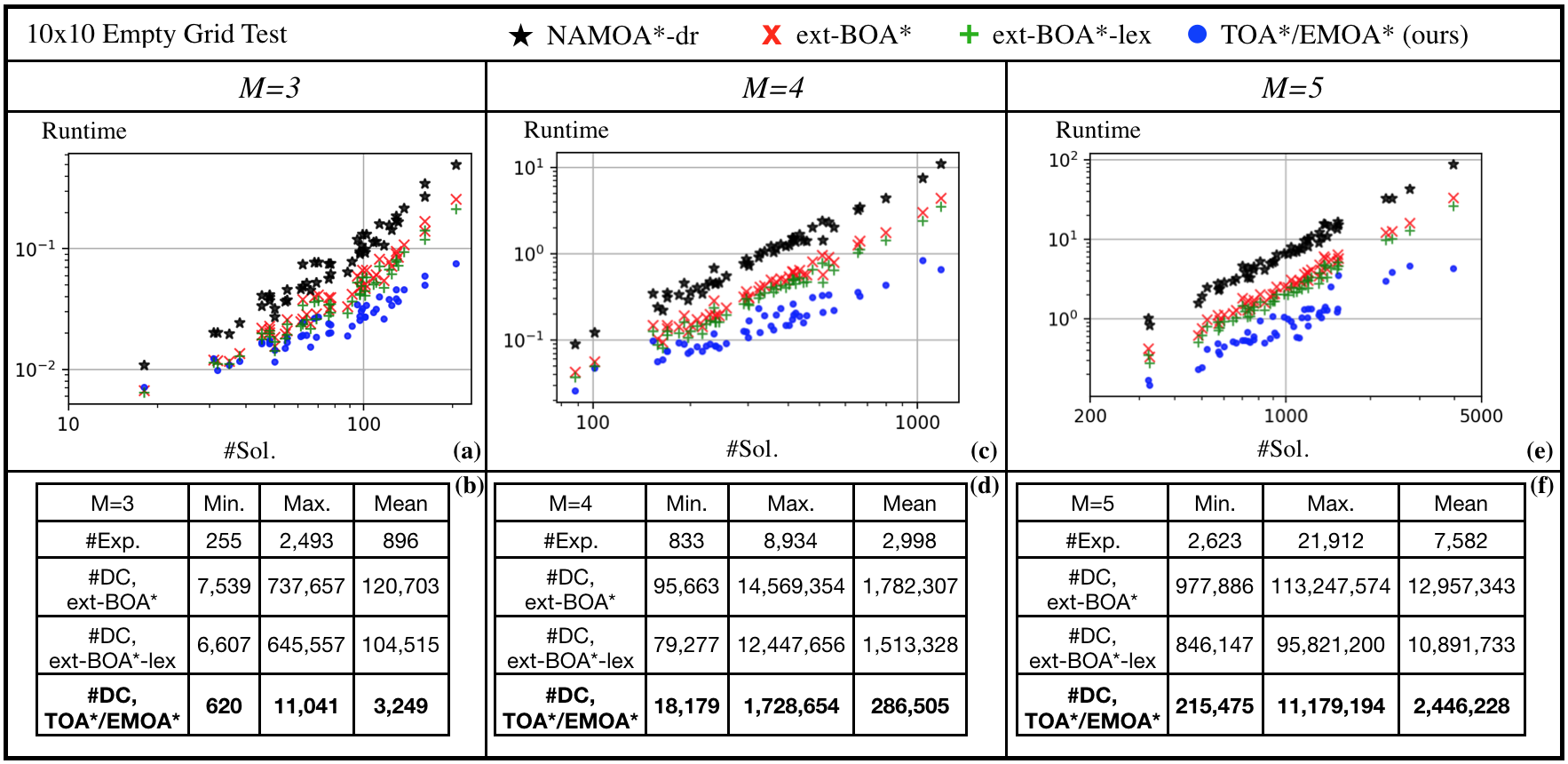}
	\caption{
		Performance comparison between TOA*/EMOA* (ours) and the baselines (NAMOA*-dr, ext-BOA*, ext-BOA*-lex) in an empty $10\times10$ map.
		The runtime (seconds) of each instance is visualized against the number of cost-unique Pareto-optimal solutions of the instance in Fig (a,c,e).
		TOA*/EMOA*, ext-BOA* and ext-BOA*-lex all follows Alg.~\ref{alg:emoa} and have the same number of expansions (\#Exp). Rows with \#DC in (b,d,f) shows the number of dominance checks required by each of the algorithms.
		To summarize, TOA*/EMOA* reduce the number of dominance checks during the search, and run up to an order of magnitude faster than the baselines. {\bf Note that the runtime axis is in log scale.}
	}
	\label{fig:M345_empty}
\end{figure*}

\subsection{Baselines and Implementation}
To verify the performance of EMOA* and TOA*, we introduce three baselines for comparison.
The \textbf{first} baseline is NAMOA*-dr~\cite{pulido2015dimensionality}, which is an algorithm in the literature that can handle an arbitrary number of objectives.

We propose a \textbf{second} baseline, which is an extension of BOA* to handle more than two objectives (hereafter referred to as ext-BOA* for simplicity).
Specifically, the \textit{FrontierCheck}, \textit{SolutionCheck} and \textit{UpdateFrontier} procedures are implemented with a naive for-loop as aforementioned in section ``The Check and Update Problems''.
Furthermore, at each node $v$, a list $Q(v)$ that consists of the non-dominated projected $\vec{g}$-vectors of the labels in the frontier set $\alpha(v)$ is introduced, and those three procedures run for-loops to conduct dominance comparisons between the projected vector of the input label and each of the project vectors in $Q(v)$.

We propose a \textbf{third} baseline, which is an ``optimized'' version of ext-BOA* and is referred to as ext-BOA*-lex, where $Q(v)$ at each node $v$ is further sorted with the lexicographic order from the minimum ($i.e.$ lex. min.) to the maximum ($i.e.$ lex. max.).
In \textit{FrontierCheck} (and \textit{SolutionCheck}), to check whether an input label $l$ is dominated or not, the procedure loops from the lex. min. to the lex. max. of $Q(v)$ (and $Q(v_d)$) and stops at the first vector in $Q(v)$ (and $Q(v_d)$) that dominates $p(\vec{g}(l))$ (and $p(\vec{f}(l))$ respectively).
Similarly, in \textit{UpdateFrontier}, to filter $Q(v)$ with an input non-dominated label $l$, the procedure first loops from the lex. max. to the lex. min. and stops at the first vector in the list that is lexicographically less than $p(\vec{g}(l))$.
Then \textit{UpdateFrontier} finds the right place to insert $p(\vec{g}(l))$ into $Q(v)$ to ensure that $Q(v)$ is still lexicographically sorted.
We call it an optimized version as we take the view that, by running a for-loop over a lexicographically sorted list, the for-loop may stop earlier, and thus the overall search is expedited.

We implement all algorithms in C++ and test on a Ubuntu 20.04 laptop with an Intel Core i7-11800H 2.40GHz CPU and 16 GB RAM without multi-threading.\footnote{Our software is at \url{https://github.com/wonderren/public_emoa}}
For a $M$-objective problem, the heuristic vectors are computed by running $M$ backwards Dijkstra search from $v_d$: the $m$-th Dijkstra search ($m=1,2,\dots, M$) uses edge cost values $c_m(e),\forall e\in E$ ($i.e.$ the $m$-th component of the cost vector $\vec{c}(e)$ of all edges).
The time to compute heuristics is negligible in comparison with the overall runtime, and we report the runtime of the algorithm that excludes the time for computing heuristics.
Finally, note that EMOA* with $M=2$ is the same as BOA*~\cite{ulloa2020simple}, which has been investigated.

\subsection{Experiment 1: Empty Map with $M=3,4,5$}

We begin by testing TOA* ($M=3$) and EMOA* ($M=4,5$) against three baselines in a small obstacle-free four-connected grid of size $10\times10$ with $v_o$ locating at the lower left corner and $v_d$ locating at the upper right corner.
Each component of the edge cost vector is randomly sampled from the integers within $[1,10]$, which follows the convention in~\cite{pulido2015dimensionality}, and 50 instances are generated.
We plot the the runtime (vertical axis) against the number of cost-unique Pareto-optimal solutions (horizontal axis) for each instance in Fig.~\ref{fig:M345_empty}.

As shown in Fig.~\ref{fig:M345_empty} (a,c,e), our TOA*/EMOA* expedites the overall search process for up to about an order of magnitude for all $M=3,4,5$ in comparison with the baselines.
Additionally, the speed-up provided by our methods becomes more obvious as the number of Pareto-optimal solutions increases, which indicates that TOA*/EMOA* are particularly advantageous when the given problem instance has numerous Pareto-optimal solutions.
Fig.~\ref{fig:M345_empty} (a,c,e) also show that ext-BOA*-lex slightly expedites the search process in comparison with ext-BOA* in general, which indicates that sorting $Q(v)$ by lexicographic order can expedite dominance checks within Alg.~\ref{alg:emoa}.
However, this expedition is negligible when comparing with our methods, which verifies the benefits of constructing balanced binary search trees to organize the frontier set at vertices.

In Fig.~\ref{fig:M345_empty} (b,d,f), Row \#Exp shows the number of expansions required by ext-BOA*, ext-BOA*-lex and TOA*/EMOA*. Note that all these three approaches follow the same workflow as shown in Alg.~\ref{alg:emoa} and thus have the same number of expansions during the search. NAMOA*-dr is omitted due to its relatively high runtime.
Rows with \#DC show the numbers of dominance checks required by the algorithms during the search.
We can observe that TOA*/EMOA* significantly reduces the number of dominance checks.
Note that \#DC serves only as a reference here, and it’s not an accurate indicator of the computational burden for the following two reasons. First, the actual implementation of dominance checks runs a for-loop over components of vectors, and this for-loop terminates without reaching the last component when non-dominance is verified. This makes each dominance check operation have varying computational efforts. Second, in TOA*/EMOA*, the component-wise scalar comparison required when traversing the balanced binary search trees is not counted as dominance checks.

\subsection{Experiment 2: Three Objectives in Various Maps}

\begin{figure}[tb]
	\centering
	\includegraphics[width=\linewidth]{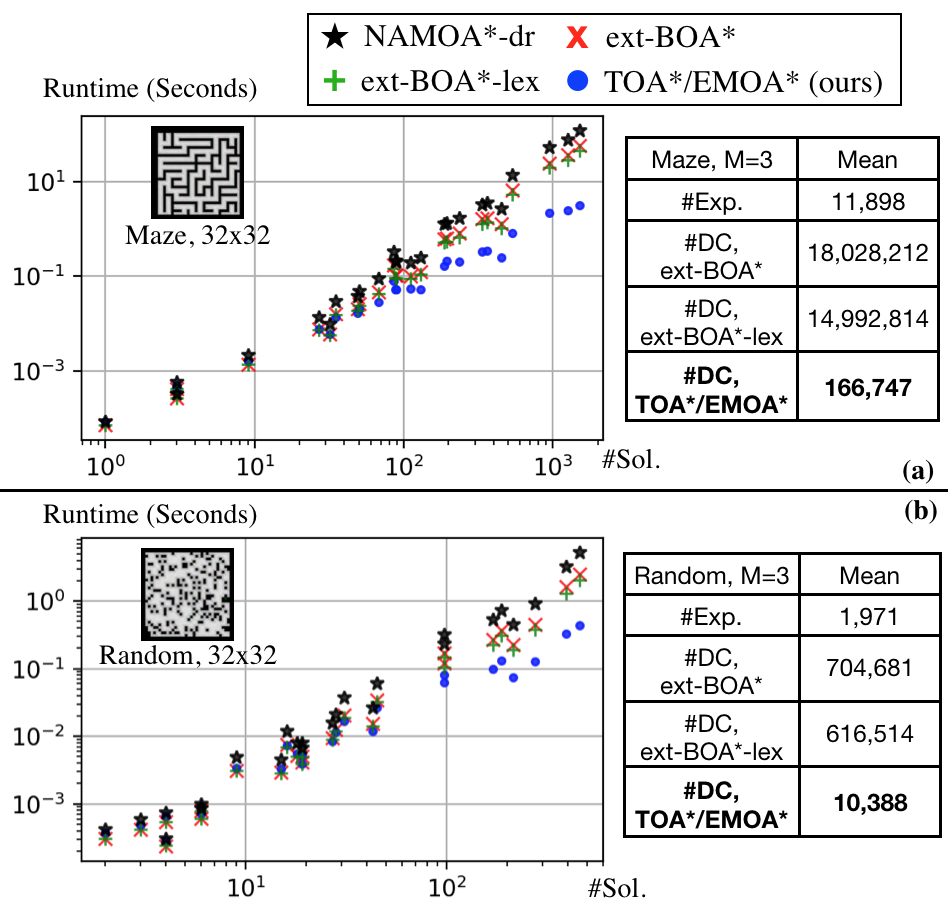}
	\caption{
		Comparison between TOA* (ours) and the baselines in various maps. The search time of each instance is visualized against the number of cost-unique Pareto-optimal solutions of the instance.
		TOA* runs up to an order of magnitude faster than the baselines due to the reduced number of dominance checks. {\bf Note that the runtime axis is in log scale.}
	}
	\label{fig:M3_maps}
\end{figure}


We then fix $M=3$ and test the algorithms in two (grid) maps selected from a online data set~\cite{stern2019multi}.
We make each grid map four-connected, and sample each component of the edge cost vector randomly from the integers within $[1,10]$.

As shown in Fig.~\ref{fig:M3_maps}, TOA* runs faster than the baselines when there are a lot of Pareto-optimal solutions (e.g. $>30$).
When the number of Pareto-optimal solutions is small, the runtime of TOA* is similar to or slower than ext-BOA*-lex.
It indicates that for problem instances with a small number of Pareto-optimal solutions, our method may not be the best choice to solve the problem.
But it's also worthwhile to note that those instances are in general not challenging, as they can be solved by either of the four approaches within 0.1 seconds.

\subsection{Experiment 3: City Road Network}

\begin{table}[htb]
	\centering
    \small
	\tabcolsep=0.2cm
	\renewcommand{\arraystretch}{1.1}
	\begin{tabular}{ | l | l | l | }
		\hline
		 & Success/All & (*) Mean/Median/Max RT
		\\ \hline
		NAMOA*-dr & 16/50 & 25.4 / 92.9 / 539.8
		\\ \hline
		ext-BOA* & 17/50 & 11.6 / 40.1 / 222.7
		\\ \hline
		ext-BOA*-lex & 17/50 & 9.7 / 33.7 / 188.8
		\\ \hline
		TOA* (ours) & \textbf{33/50} & \textbf{1.8} / \textbf{5.0} / \textbf{31.0}
		\\ \hline
	\end{tabular}
	\caption{This table shows number of succeeded instances in New York City map from a online data set.
	Symbol (*) means the mean/median/max runtime (RT) are taken over the 16 instances where all four algorithms succeed.
	The mean, median and maximum number of Pareto-optimal solutions for those 16 instances are 389, 327 and 1061 respectively. The minimum runtime is omitted as there exists a trivial instance with only one Pareto-optimal solution and all algorithms terminate within a few micro-seconds.
	Our TOA* doubles the number of succeeded instances within a limited runtime of 600 seconds. Over the 16 solved instances, TOA* requires about 1/6 of the runtime of ext-BOA*-lex.}
	\label{tab:NY3}
\end{table}

Finally, we evaluate TOA* and the baselines in the New York City map (a graph with 264,346 vertices and 733,846 edges) from a online data set.\footnote{\url{http://www.diag.uniroma1.it//~challenge9/download.shtml}}
This data set provides distance ($c_1$) and travel time ($c_2$) for each edge. We introduce a third type of cost as follows (which is deterministic and reproducible).
Let $deg(v)$ denote the degree (number of adjacent vertices) of $v\in V$, and let $deg(e) := \frac{deg(u)+deg(v)}{2}, e=(u,v)\in E$.
If $deg(e) \geq 4$, $c_3(e) = 2$, otherwise $c_3(e)=1$.
The design of $c_3$ is motivated by hazardous material transportation~\cite{erkut2007hazardous}, where the transportation over busy edges can lead to higher risk if leakage happens, and $deg(e)$ is an indicator about how busy an edge is.
We discuss the results in the caption of Table~\ref{tab:NY3} and provide an illustration of the Pareto-optimal solutions in Fig.~\ref{EMOA:fig:NY_pic} in the Appendix.

%% file: related.tex
MO-SPP algorithms range from exact approaches~\cite{moastar,mandow2008multiobjective,ulloa2020simple} to approximation methods~\cite{goldin2021approximate,perny2008near,warburton1987approximation}, trading off solution optimality for computational efficiency.
This work belongs to the category of exact approaches.



Another related work is the Kung's method~\cite{kung1975finding} which addresses the following problem:
Given an arbitrary set $A$ of $M$ dimensional vectors ($M\geq2$), compute $ND(A)$, the non-dominated subset of $A$.
The DC and NSU problems introduced in this work as mentioned in Def.~\ref{problem:check} and \ref{problem:update} can be regarded as \emph{incremental} versions of the problem solved by Kung's method, since the frontier set is constructed in an incremental manner during the MOA*-like search.

%% file: conclude.tex
This work considers a Multi-Objective Shortest Path Problem (MO-SPP) with an arbitrary number of objectives.
In this work, we observe that, during the search process of MOA*-like algorithms, the frontier set at each vertex is computed incrementally by solving the Dominance Check (DC) problem and Non-Dominated Set Update (NSU) problems iteratively.
Based on this observation, we develop a balanced binary search tree (BBST)-based approach to efficiently solve the DC and NSU problems in the presence of an arbitrary number of objectives.
With the help of the BBST-based methods and the existing fast dominance check techniques, we develop the Enhanced Multi-Objective A* (EMOA*), which computes all cost-unique Pareto-optimal paths for MO-SPP problems.
EMOA* is a generalization of BOA*.
We also develop the TOA*, an improved version of EMOA* when there are three objectives.
We discuss the correctness and the computational complexity of the proposed methods, and verify them with massive tests.
The numerical result shows that TOA* and EMOA* runs up to an order of magnitude faster than all the baselines, and are of particular advantage for problems with a large number of Pareto-optimal solutions.

%% file: appendix.tex
\newtheorem{innercustomthm}{Theorem}
\newenvironment{customthm}[1]
  {\renewcommand\theinnercustomthm{#1}\innercustomthm}
  {\endinnercustomthm}
  
  \newtheorem{innercustomlem}{Lemma}
\newenvironment{customlem}[1]
  {\renewcommand\theinnercustomlem{#1}\innercustomlem}
  {\endinnercustomlem}
  
  \newtheorem{innercustomcol}{Corollary}
\newenvironment{customcol}[1]
  {\renewcommand\theinnercustomcol{#1}\innercustomcol}
  {\endinnercustomcol}

\section{Appendix}
\subsection{Proof of TOA*}

\begin{customthm}{1}
	Given $b$, a two dimensional vector, and an arbitrary node $n$ in $\mathcal{T}_B$, if (i) $n.key \nleq b$ and (ii) $b >_{lex} n.key$, then the key of any nodes in the left sub-tree of $n$ cannot dominate $b$.
\end{customthm}

\begin{proof}
    Let subscript denote the specific component within a vector.
	From (i) and (ii), we have $b_1 \geq n.key_1$ and $b_2 < n.key_2$.
	For any node $n'$ in the left sub-tree of $n$, by construction of the tree, $n' <_{lex} n$ and thus $n'.key_1 \leq n.key_1$.
	Additionally, by definition, the key of every pair of nodes in $\mathcal{T}_B$ are non-dominated and non-equal to each other, thus $n'.key_2 > n.key_2$.
	Put them together, we have $$b_2 < n.key_2 < n'.key_2.$$
	Thus, $b$ is not dominated by $n'.key$. Since $n'$ can be any nodes in the left sub-tree of $n$, the theorem is proved.
\end{proof}

\subsection{Proof of EMOA*}
	
During the search process, at line 5 of Alg.~\ref{alg:emoa}, we say a label is \emph{extracted} from OPEN.
At line 11 of Alg.~\ref{alg:emoa}, we say a label is \emph{expanded}.
Clearly, the set of expanded labels during the search is a subset of extracted labels.
Additionally, at line 12 of Alg.~\ref{alg:emoa}, we say a new label is \emph{generated}.

\begin{customlem}{1}
	When expanding a label $l$, each successor label $l'$ has $f_1$ value no smaller than the $f_1$ value of $l$.
\end{customlem}

\begin{proof}
	As $\vec{h}$-vectors (of labels) are consistent, $c_1(v(l),v(l')) + h_1(v(l')) \geq h_1(v(l))$.
	Therefore, $f_1(l') = g_1(l') + h_1(v(l')) = g_1(l) + c_1(v(l),v(l')) + h_1(v(l')) \geq g_1(l) + h_1(v(l)) = f_1(l)$.
\end{proof}

\begin{customlem}{2}
	During the search of Alg.~\ref{alg:emoa}, the sequence of extracted labels from OPEN has non-decreasing $f_1$ values.
\end{customlem}

\begin{proof}
	In each search iteration, EMOA* extracts the label $l$ with the lexicographically minimum $\vec{f}$, which means $l$ has the minimum $f_1$ value among any other label $l'$ in OPEN.
	With Lemma~\ref{lem:non_decreasing_f1_successor}, any successor label $l''$ to be generated after the expansion of $l$ (or $l'$) has no smaller $f_1$ value than the $f_1$ value of $l$ (or $l'$).
	Therefore, the sequence of extracted labels from OPEN has non-decreasing $f_1$ values.
\end{proof}

\begin{customcol}{1}
	During the search, the sequence of extracted and expanded labels at a specific vertex has non-decreasing $f_1$ and $g_1$ values.
\end{customcol}
As the sequence of extracted labels at a specific vertex is a subset of all labels that are extracted from OPEN,
this corollary is obvious given Lemma~\ref{lem:non_decrease_f1}. 

\begin{customcol}{2}
	During the search, for a label $l$ that is extracted from OPEN, we have $\vec{g}(l')\leq \vec{g}(l), l' \in\alpha(v(l))$ if and only if $p(\vec{g}(l')) \leq p(\vec{g}(l)), l' \in\alpha(v(l))$.
\end{customcol}


\begin{customthm}{2}
	EMOA* computes a maximal set of cost-unique Pareto-optimal paths connecting $v_o$ and $v_d$ at termination.
\end{customthm}

\begin{proof}
In each search iteration, Alg.~\ref{alg:emoa} extracts a label $l$ from OPEN (line 5), whose $\vec{f}$-vector is the lexicographical minimum in OPEN.
It means none of the remaining labels in OPEN can dominate $l$.
As procedures \textit{FrontierCheck} and \textit{SolutionCheck} are correct: label $l$ is discarded if and only if it is dominated by or equal to some other expanded labels, which means it can not lead to a cost-unique Pareto-optimal solution (line 6-7).
If label $l$ is not discarded, it is then added to the frontier set $\alpha(v(l))$ (line 8), which ensures that $\alpha(v(l))$ contains cost-unique non-dominated labels.
When Alg.~\ref{alg:emoa} terminates, each of the labels in $\alpha(v_d)$ must represent a cost-unique Pareto-optimal solution.
Finally, during the expansion, all possible successor labels are generated and the non-dominated ones are inserted into OPEN for future expansion.
The algorithm terminates only when all labels are either expanded or discarded, which guarantees that all cost-unique Pareto-optimal solutions are found.
\end{proof}


\subsection{Experiment 3: City Road Network Illustration}
\begin{figure*}[h!]
	\centering
	\includegraphics[width=\linewidth]{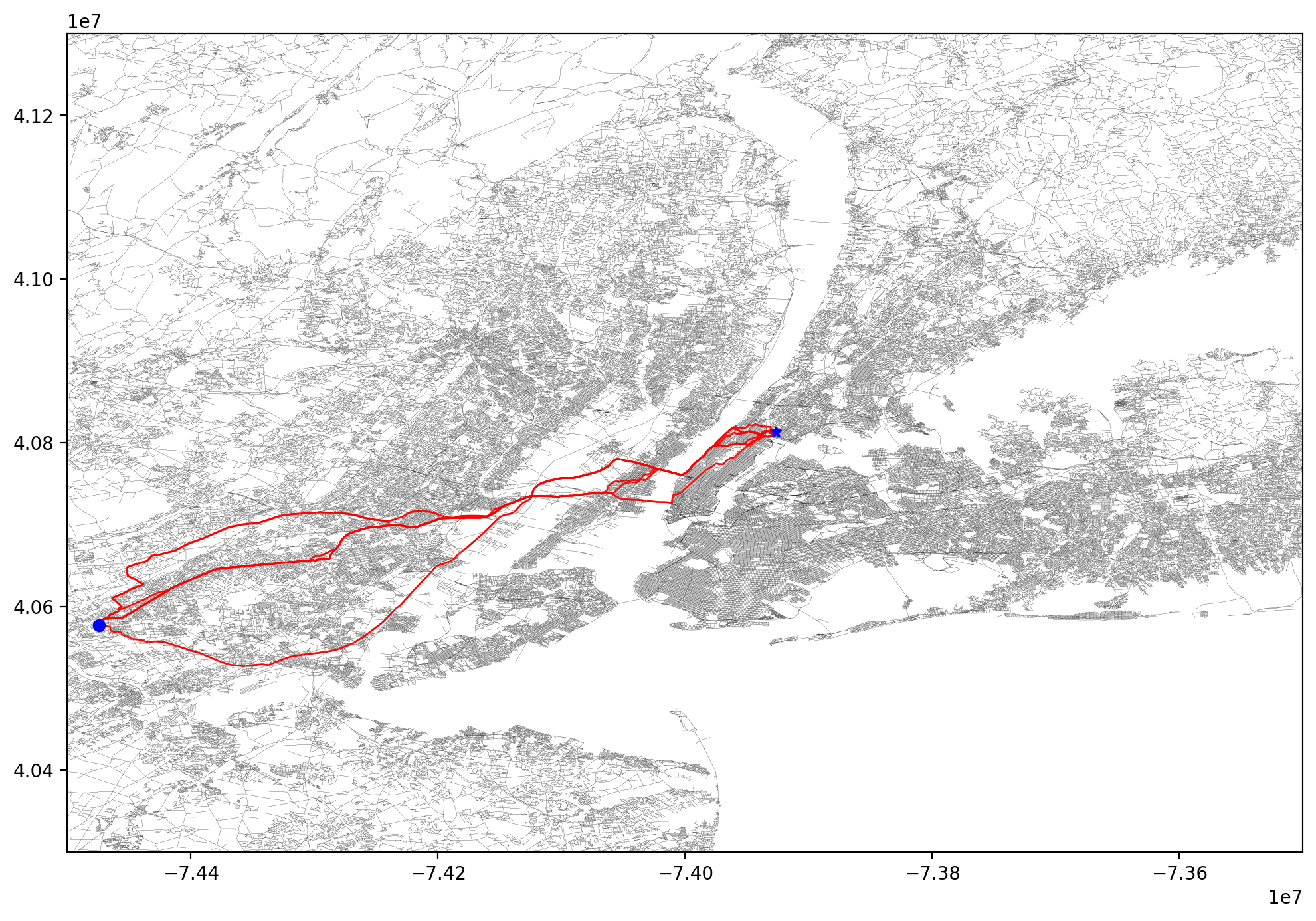}
	\caption{An illustration of a subset of the computed Pareto-optimal solution paths for a problem instance in the New York City map.
	}
	\label{EMOA:fig:NY_pic}
\end{figure*}

A subset of the computed Pareto-optimal solution paths for a problem instance in the New York City map (with 264,346 vertices and 733,846 edges) is shown in Fig.~\ref{EMOA:fig:NY_pic}.
